\documentclass[default,iicol]{sn-jnl}

\usepackage{subcaption}
\usepackage{graphicx}%
\usepackage{multirow}%
\usepackage{amsmath,amssymb,amsfonts}%
\usepackage{amsthm}%
\usepackage{mathrsfs}%
\usepackage[title]{appendix}%
\usepackage{xcolor}%
\usepackage{textcomp}%
\usepackage{manyfoot}%
\usepackage{booktabs}%
\usepackage{algorithm}%
\usepackage{algorithmicx}%
\usepackage{algpseudocode}%
\usepackage{listings}%
\usepackage{tikz}
\usetikzlibrary{shapes.geometric}
\usepackage{adjustbox}
\usepackage{academicons}
\usepackage{hyperref}

\begin{document}
\title{Unsupervised out-of-distribution detection by restoring lossy inputs with variational autoencoder}

\author[]{\fnm{Zezhen} \sur{Zeng}}\email{zengzz@zhejianglab.com}
\author*[]{\fnm{Bin} \sur{Liu}}\email{bins@ieee.org}

\affil[]{\orgdiv{Research Center for Applied Mathematics and Machine Intelligence}, \orgname{Zhejiang Lab}, \orgaddress{\street{Hangzhou}, \city{Hangzhou}, \postcode{311000}, \country{China}}}

\abstract{Deep generative models have been demonstrated as problematic in the unsupervised out-of-distribution (OOD) detection task, where they tend to assign higher likelihoods to OOD samples. Previous studies on this issue are usually not applicable to the Variational Autoencoder (VAE). As a popular subclass of generative models, the VAE can be effective with a relatively smaller model size and be more stable and faster in training and inference, which can be more advantageous in real-world applications. In this paper, We propose a novel VAE-based score called Error Reduction (ER) for OOD detection, which is based on a VAE that takes a lossy version of the training set as inputs and the original set as targets. Experiments are carried out on various datasets to show the effectiveness of our method, we also present the effect of design choices with ablation experiments. Our code is available at: https://github.com/ZJLAB-AMMI/VAE4OOD.}

\keywords{out-of-distribution, variational autoencoder}

%

\maketitle

\section{Introduction}
The safety and reliability of deep learning models in real-world applications are crucial for modern society. One essential aspect of ensuring model reliability is the capability to differentiate between in-distribution (ID) data and out-of-distribution (OOD) data. Several studies have explored using classifiers for OOD detection \cite{hendrycks2016baseline, lakshminarayanan2017simple, liang2017enhancing, hsu2020generalized, sun2021react}. These models are typically trained to learn category-based features and can easily differentiate ID and OOD samples. Although this line of work has shown promising results, it necessitates model supervision and is hence not feasible in scenarios where category labels are unavailable.

In real-world applications, obtaining data labels may not always be practical. In such cases, unsupervised deep generative models are advantageous in detecting OOD samples. Unlike supervised models, these models do not require labelled data for training and can learn to detect OOD samples by modelling the underlying distribution of ID samples \cite{kingma2013auto, oord2016pixel, salimans2017pixelcnn++, kingma2018glow, maaloe2019biva}. A well-trained generative model is expected to assign higher likelihoods to ID samples and lower likelihoods to OOD samples as the latter originate from distributions that differ from the distribution of ID samples \cite{bishop1994novelty}.

Contrary to theoretical expectations, several studies have found that generative models often assign higher likelihoods to OOD samples than to ID samples \cite{nalisnick2018deep}. For example, the generative model trained on the CIFAR10 can assign higher likelihoods on SVHN samples than on CIFAR10 samples. This counterintuitive phenomenon has spurred a series of studies aimed at understanding and resolving the likelihood misalignment issue \cite{ren2019likelihood, serra2019input, xiao2020likelihood, schirrmeister2020understanding, havtorn2021hierarchical, zhang2021understanding, li2022out, flototilted}. However, most of them result in performance degeneration when the model is set to be Variational Autoencoders (VAE) \cite{kingma2013auto, Rezende:2014:SBA:3044805.3045035}. \cite{xiao2020likelihood} claim that VAEs have less separated likelihoods for ID and OOD samples, which makes OOD detection more challenging for VAEs.  Compared to other likelihood-based generative models such as PixelCNN \cite{salimans2017pixelcnn++} and Glow\cite{kingma2018glow}, VAE is faster in training and can be effective with a relatively smaller model architecture, which can be important in memory-restricted cases. Thus, an efficient VAE-based OOD detection method is needed.

In this paper, we introduce a new OOD score based on VAEs called Error Reduction (ER). By setting the input as a lossy version of the training set and maintaining the original training set as the reconstruction target, the VAE can perform as an image restoration model. ER measures the difference between two errors. One error is between the lossy sample and the corresponding original version, while the other error is the reconstruction error of the VAE. ER value essentially serves as a gauge of improvement, as the reconstruction should be in closer proximity to the original image in comparison to the lossy input. The process of obtaining a lossy input is intended to bridge the gap between ID samples and OOD samples. This approach encourages the model to treat all inputs as emanating from the same distribution while keeping outputs close to the ID dataset, which results in a better improvement for ID samples. Our proposed method consistently performs well across various OOD datasets without the need for model fine-tuning during the test phase. Experimental results demonstrate that the proposed ER score achieves comparable or better results than other VAE-based scores on a wide range of datasets.


\section{Related Work}
Although assigning high likelihoods to OOD samples is a common issue with deep generative models in OOD detection, some specific generative models can evade the likelihood misalignment curse \cite{maaloe2019biva, grathwohl2019your, havtorn2021hierarchical}. The energy-based model can be trained in both supervised and unsupervised modes \cite{grathwohl2019your}. Recent works on the hierarchical VAE \cite{maaloe2019biva, havtorn2021hierarchical,li2022out} demonstrate that hierarchical VAEs can circumvent the likelihood misalignment issue based on the likelihood ratio between different levels of latent variables.

Several studies attempt to address the likelihood misalignment problem using vanilla likelihood-based generative models. Likelihood Ratio assumes an input can be separated into a common background component and a data-specific semantic component \cite{ren2019likelihood}. One model captures only the background information, while another is trained in the standard setting, and the Likelihood Ratio is computed between the two models. 
Input Complexity (IC) \cite{serra2019input} is based on the observation of a negative correlation between the generative models' likelihoods and the complexity estimates. They suggest an OOD score that compensates for the likelihood with the complexity estimation. 
However, these methods do not perform well with VAEs. The Likelihood Regret (LR) \cite{xiao2020likelihood} is the first effective OOD score for VAEs that uses the ratio between a VAE trained on the whole dataset and a VAE optimised with a single test sample. But LR needs fine-tuning for every single test sample which is time-consuming.
FRL \cite{cai2023out} adds a channel of high-frequency information into training to improve OOD detection results. Tilted VAE \cite{flototilted} introduced an exponentially tilted prior distribution to address the likelihood misalignment issue. However, this method's performance does not generalise well to different datasets. Compared to these methods, our method follows the training strategy of the vanilla VAE while the only modification is using lossy inputs rather than the original inputs. Our model also does not ask for further fine-tuning.

In addition to the likelihood, the reconstruction error is another commonly used metric to address the OOD problem. However, the reconstruction error also suffers from misalignment between the reconstruction errors of ID and OOD samples. To alleviate this issue, \cite{denouden2018improving} propose using the Mahalanobis distance between the test sample and the mean of the training set as a regularisation term for the reconstruction error.

\section{Background}
In this section, we first introduce the problem formulation of OOD detection with generative models and then introduce the VAE.
\subsection{Problem formulation}
 Assuming $\boldsymbol{X}$ = $\{\boldsymbol{x}_{i}\}^N_{i=1}$ is a set of $N$ i.i.d. samples drawn from an in-distribution $\boldsymbol{x} \sim p(\boldsymbol{x})$, the objective is to detect whether an unobserved sample is an ID sample or not. A generative model trained on $\boldsymbol{X}$ is expected to learn a good approximation to the underlying distribution $p(\boldsymbol{x})$. Consequently, OOD samples should be assigned lower likelihood under a well-trained generative model. One approach is to set a threshold based on the validation set and consider a test sample as an OOD sample if its likelihood is lower than the threshold. However, using the likelihood directly can be problematic when the test ID/OOD pairs are like FashionMNIST/MNIST and CIFAR10/SVHN \cite{nalisnick2018deep,ren2019likelihood,serra2019input}.

 \subsection{Variational autoencoder}
Building on the previous notes, we define a latent variable $\boldsymbol{z}$ that is sampled from a prior distribution $p(\boldsymbol{z})$, usually a standard Gaussian distribution.
As an important type of deep generative model, the VAE aims to maximize the marginal likelihood $p_{\theta} = \int_{z}{p_{\theta}(\boldsymbol{x}|\boldsymbol{z})p(\boldsymbol{z})d\boldsymbol{z}}$ with some unknown parameters $\theta$. However, the likelihood is typically intractable in high-dimensional spaces. Therefore, the VAE uses an inference model $q_{\phi}(\cdot)$, commonly known as the encoder, to approximate the true posterior $p_{\theta}(\boldsymbol{x}|\boldsymbol{z})$, where $\phi$ is the parameters of the inference model. This leads to an objective function known as the Evidence Lower Bound (ELBO):
\begin{equation}
\begin{aligned}
  &\log p_{\theta}(\boldsymbol{x}) \\
  &\geq
   \mathbb{E}_{q_{\phi}(\boldsymbol{z}|\boldsymbol{x})}[\log p_{\theta}(\boldsymbol{x}|\boldsymbol{z})]
  - D_{KL}(q_{\phi}(\boldsymbol{z}|\boldsymbol{x})||p(\boldsymbol{z}))
\end{aligned}\label{eq:elbo}
\end{equation}

where $q_{\phi}(\boldsymbol{z}|\boldsymbol{x})$ is the approximation to the true posterior, and $D_{KL}(\cdot||\cdot)$ is the KL divergence. The first term is the reconstruction loss, and the second term is the KL divergence between the approximate posterior and the prior distribution.

\section{Method}\label{4.2}
The idea of ER is based on an analysis of LR \cite{xiao2020likelihood}. LR is a VAE-based OOD score that measures the improvement of the likelihood of a test sample from the VAE optimised with the single test sample over the VAE trained with training sets. According to the experiments in \cite{xiao2020likelihood}, LR can perform well only if the decoder is frozen. Here we analyse this phenomenon from the learning preference of the VAE.

From the perspective of Bits-Back Coding \cite{frey1996free}, the objective of optimising the VAE is to minimise the code length needed to transmit $\boldsymbol{x}$, the code length is determined by:

\begin{align}
  \mathcal{L} &=
   \mathbb{E}_{\boldsymbol{x}, \boldsymbol{z}}[\log q(\boldsymbol{z}|\boldsymbol{x})-\log p(\boldsymbol{z})-\log p(\boldsymbol{x}|\boldsymbol{z})] \nonumber\\
  &=\mathbb{E}_{\boldsymbol{x}}[-\log p(\boldsymbol{x}) + D_{KL}(q(\boldsymbol{z}|\boldsymbol{x})||p(\boldsymbol{z}|\boldsymbol{x}))].
\end{align}

where $\mathbb{E}_{\boldsymbol{x}, \boldsymbol{z}}$ denotes $\mathbb{E}_{\boldsymbol{x}\sim\boldsymbol{X}, \boldsymbol{z}\sim q(\boldsymbol{z}|\boldsymbol{x})}$ and $\mathbb{E}_{\boldsymbol{x}}$ denotes $\mathbb{E}_{\boldsymbol{x}\sim\boldsymbol{X}}$. To minimise the $\mathcal{L}$, the decoder prefers to model the training data alone without using the information in $\boldsymbol{z}$, in which case the extra cost $D_{KL}(q(\boldsymbol{z}|\boldsymbol{x})||p(\boldsymbol{z}|\boldsymbol{x}))$ can be avoided \cite{chen2016variational,bowman2015generating}. In other words, the VAE prefers to model the information within the decoder during the training procedure.

Once we understand the preference of the VAE, it is easy to explain why LR performs worse with an optimised decoder. There is a trend that the decoder models a relatively large portion of information of training set locally. LR measures the change in likelihood when we optimise the encoder of the VAE with a test sample. Thus, the magnitude of the change of each ID sample is constrained since most of the information is modelled by the decoder. In contrast, there is less information of OOD samples learned by either the encoder or the decoder, so the OOD samples can have a higher value of LR compared to the ID samples if the VAE is only optimised with the encoder. This gap can be reduced when the decoder is also optimised, resulting in worse OOD detection performance.

We also show the difference between optimising only the encoder and optimising only the decoder for a pre-trained VAE qualitatively in Fig.~\ref{tab:mdl}. The pre-trained VAE is trained on the original FashionMNIST \cite{xiao2017fashion} and we show the reconstructions for both FashionMNIST and MNIST \cite{lecun1998mnist}. We optimise the encoder or decoder with the lossy version of the entire dataset for 50 epochs. Here, we denote the lossy version of sample $\boldsymbol{x}$ as $\boldsymbol{x}_{lossy}$. There are various ways to obtain $\boldsymbol{x}_{lossy}$, such as mask operation and downsampling followed by upsampling. We use nearest neighbour interpolation as the sampling method to obtain $\boldsymbol{x}_{lossy}$. In Fig.~\ref{tab:mdl}, all reconstructions are generated by feeding $\boldsymbol{x}_{lossy}$. The VAE with an optimised encoder can produce similar reconstructions to the original VAE, while the VAE with an optimised decoder can generate reconstructions closer to the lossy inputs, indicating that the decoder stores more information than the latent variables learned by the encoder.


\begin{table*}[h]
\begin{tabular}{cll}
\raisebox{2.5mm} {Raw images}    & \includegraphics[scale=0.65]{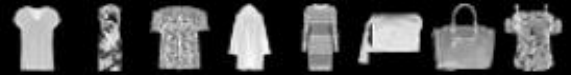} & \includegraphics[scale=0.65]{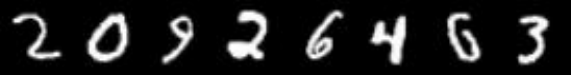} \\
\raisebox{2.5mm} {Lossy images}     & \includegraphics[scale=0.65]{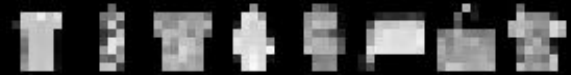}  & \includegraphics[scale=0.65]{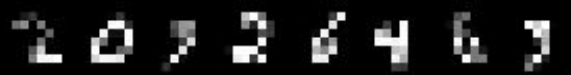} \\
\raisebox{2.5mm} {Original VAE} &\includegraphics[scale=0.65]{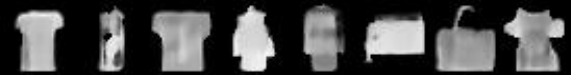}  & \includegraphics[scale=0.65]{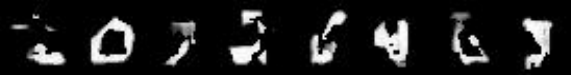}   \\
\raisebox{2.5mm}{Opt encoder} & \includegraphics[scale=0.65]{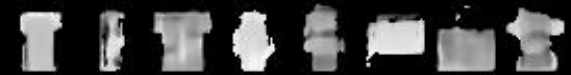} & \includegraphics[scale=0.65]{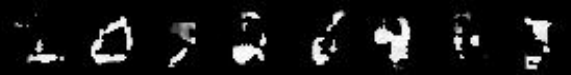} \\
\raisebox{2.5mm}{Opt decoder} & \includegraphics[scale=0.65]{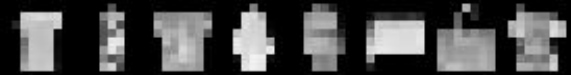}& \includegraphics[scale=0.65]{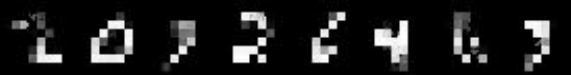}
\end{tabular}
\captionof{figure}{The first row is the original image, and the second row is the lossy version of the first row. The subsequent rows show the corresponding reconstructions of lossy images. ``Original VAE" refers to a pre-trained VAE trained on the raw images. ``Opt encoder" refers to the pre-trained VAE with an optimised encoder, and ``Opt decoder" refers to the pre-trained VAE with an optimised decoder.}
\label{tab:mdl}
\end{table*}

\subsection{Error reduction}
 Based on the VAE's training preference, we propose our OOD detection score called Error Reduction (ER), which is computed by utilising a VAE trained to reconstruct training data by feeding a lossy version of the training data. ER quantifies the improvement achieved when transitioning from the lossy version to the reconstructed images. Since the desired VAE is designed to reconstruct an image from a lossy version of that image, the decoder is predisposed to model the information that can contribute more to reconstruction, such as textures or semantic features prevalent in the dataset. Concurrently, the lossy transformation used for inputs serves to diminish the distinction between ID and OOD samples during the test phase. If we consider an entirely black image as the origin of all kinds of images, and a dataset as a collection of images located far away from the origin with an average distance $\boldsymbol{l}$. We define $\boldsymbol{l}_{in}$ and $\boldsymbol{l}_{ood}$ as the average distance of the ID samples and the OOD samples, respectively. For simplicity, the OOD samples described here are from the same OOD dataset. We assume that the ratio of information loss after the lossy operation is close to a constant $\epsilon$ where $\epsilon$ is limited to being less than or equal to 1, as the procedure of obtaining the lossy version of images is the same for all images. Thus the remaining distance should be $(1 - \epsilon) \boldsymbol{l}_{in}$ and $(1 - \epsilon) \boldsymbol{l}_{ood}$, where a higher value of $\epsilon$ makes ID and OOD samples closer. For the extreme case of $\epsilon = 1$, all images converge to the origin, resulting in a pure black image. As input samples get closer after the lossy operation, the corresponding latent variables derived from the encoder also converge. Meanwhile, as the decoder is biased towards ID samples, it is expected to reconstruct OOD samples in the style of ID samples since the latent variables are close, which implies that the improvement in image restoration for OOD samples might be less than for ID samples.

We first define a simple metric to measure the improvement:
\begin{align}
    E(\boldsymbol{x}) = e(x, x_{lossy}) - e(x, x^{\prime}), \label{eq:xx}
\end{align}

The metric $E(\boldsymbol{x})$ is defined as the difference between two errors, where the first term is the error between the original image $\boldsymbol{x}$ and the lossy version $\boldsymbol{x}_{lossy}$. The second term is the error between the input image $\boldsymbol{x}$ and the corresponding reconstruction $\boldsymbol{x}^{\prime}$ from the VAE. We choose the reconstruction error term of the VAEs' objective function as the default error function and we also test the performance of different choices in Section~\ref{se:ex}. Since the model is optimised to restore the lossy version of the training set, the second term is supposed to be lower than the first term. A higher value of $E(\boldsymbol{x})$ can indicate a better improvement in reconstruction quality.

However, using Equation~\ref{eq:xx} as the OOD detection score directly can be problematic. Because both $e(x, x_{lossy})$ and $e(x, x^{\prime})$ are non-negative, which makes $E(\boldsymbol{x}) \leq e(x, x_{lossy})$ and $e(x, x_{lossy})$ can be considered as an upper bound of $E(\boldsymbol{x})$. Moreover, the value of $e(x, x_{lossy})$ depends on the data distribution and OOD samples can have a larger value than ID samples, resulting in a higher upper bound for the $E(\boldsymbol{x})$ values of OOD samples. Consequently, even if there is only a marginal degree of improvement, the OOD samples can result in a higher value of $E(\boldsymbol{x})$ compared to ID samples. To address this issue, we propose using the estimated input complexity to compensate for the error, which results in the following expression:
\begin{align}
    ER(\boldsymbol{x}) = &e(x, x_{lossy}) - e(x, x^{\prime}) \nonumber\\
    & + \lambda (L(\boldsymbol{x}) - L(\boldsymbol{x}_{lossy})).
\end{align}

where $L(\cdot)$ is an estimate of the Kolmogorov complexity \cite{kolmogorov1963tables} and $\lambda$ is a weight parameter. Intuitively, imagine a pure black image as the ultimate destination of $\boldsymbol{x}_{lossy}$ and a pure white image as the original image is supposed to give the highest $e(x, x_{lossy})$ with pixel-based error function. Meanwhile, a fully monochrome image has the lowest complexity since all pixel values are the same. Thus, adding image complexity as a regularisation term can alleviate the issue of Equation~\ref{eq:xx} that OOD samples might yield higher values.

In our paper, we use a Portable Network Graphics (PNG)-based image compressor, as suggested in IC \cite{serra2019input}. The term $L(\boldsymbol{x}) - L(\boldsymbol{x}_{lossy})$ represents the complexity change from the raw sample to the lossy sample. Note that the complexity change is not necessary to be added if $E < 0$, as $E$ is expected to be positive to represent an improvement for test samples.

\section{Experiments}\label{se:ex}
In this section, we evaluate the effectiveness of our method on various datasets. We first describe the experimental setup and then demonstrate the performance of our method, further ablation studies are provided at the end.

\subsection{Experimental setup}\label{setup}
We present the datasets, model architecture, implementing details and the evaluation metric in detail in the following.
\subsubsection{Datasets}
We demonstrate our method on a widely used ID/OOD dataset pair: CIFAR10 \cite{krizhevsky2009learning}/SVHN \cite{netzer2011reading}. To better demonstrate the generalization ability of our method, we also introduce additional OOD datasets: FashionMNIST \cite{xiao2017fashion}, MNIST \cite{lecun1998mnist}, KMNIST \cite{clanuwat2018deep}, CelebA \cite{liu2015faceattributes}, LSUN \cite{yu2015lsun}, and two synthetic datasets, Noise and Constant in our evaluation \cite{xiao2020likelihood}, all images are resized to 32$\times$32. The Noise dataset are randomly sampled from a uniform distribution in the range [0, 255] for each pixel, while the Constant dataset consists of images where all pixels in a single image have the same value, sampled from the same uniform distribution in the range [0, 255]. All images in these datasets are resized to 32$\times$32. When the OOD samples are colourful datasets and the model is trained on FashionMNIST, we use the first channel to conduct the OOD test. When the model is trained on CIFAR10, we stack three copies of grayscale images. We use 5000 random images to compute the AUROC.

In addition, we follow the setting in FRL \cite{cai2023out}. We evaluate our method on high-resolution datasets CelebA with four OOD datasets: iNaturalist\cite{inaturalist21}, Places \cite{zhou2017places}, SUN \cite{SUN} and Textures \cite{cimpoi14describing}, all images are resized to 128$\times$128.

\begin{table*}[ht]
\centering
\caption{Details of the model architecture. The parameters of Conv and Deconv are output channels, kernel size and stride. $b$ is the batch size, $z$ is the latent dimension size, $w$ is the width of images and $c$ is the final output channel. }
\label{tab:model}
\begin{tabular}{cc}
\\
\toprule
  Encoder    & Decoder               \\
\midrule
   Conv($f$, 5, 1), LeakyReLU   & Deconv(2$f$, $w$/4, 1), LeakyReLU                \\
   Conv($f$, 5, 2), LeakyReLU    & Deconv(2$f$, 5, 1), LeakyReLU               \\
   Conv(2$f$, 5, 1), LeakyReLU    & Deconv($f$, 5, 2), LeakyReLU         \\
   Conv(2$f$, 5, 2), LeakyReLU    & Deconv($f$, 5, 1), LeakyReLU                \\
   Reshape to $b$ $\times$ $z^\prime$    & Deconv($f$, 5, 2), LeakyReLU            \\
   2*Linear($z^\prime$, z) for $\mu$ and $\sigma$     & Conv($c$, 5, 1)        \\
\bottomrule
\end{tabular}\hfill%
\end{table*}

\subsubsection{Model architecture}
The backbone of our VAE is mainly based on \cite{flototilted}. Details are shown in Table~\ref{tab:model}. The $f$ is 32 for CIFAR10 and 64 for CelebA respectively.

\subsubsection{Implementing details}
We use the ADAM optimiser \cite{kingma2014adam} with a learning rate of 3e-4 for models, and the batch size $b$ is 128. The latent dimension size $z$ is 100 for CIFAR10 and CelebA. The training epoch is 200 for CIFAR10 and CelebA. The output channel $c$ is parameterised by a factorial 256-way categorical distribution as described in \cite{xiao2020likelihood}.

\subsubsection{Evalution metric}
For method evaluation, We adopt the threshold-free evaluation metric, Area Under the Receiver Operator Characteristic (AUROC), as the primary metric. Higher AUROC indicates better performance. All the reported results are averaged over 5 runs.

\subsection{Results}\label{comparison}
 Following the setting described above, we present the experimental results and compare our method with other VAE-based methods \cite{kingma2013auto,ren2019likelihood, serra2019input, xiao2020likelihood, flototilted, cai2023out}. In the comparison, we use the nearest neighbour interpolation with the scale factor set to 4 to get the lossy inputs. We show the results in Table~\ref{tab:compare}. Also, We notice that the original implementation of Tilt + WIM requires batches of OOD samples from the same dataset to optimise the model, and the model needs to be optimised for different OOD datasets respectively. However, it is unrealistic to collect batches of OOD samples for all possible OOD datasets in real applications, which makes the original implementation of Tilt + WIM unfair to other models. Thus, we implement the method based on their released code by replacing batches of OOD samples with one single OOD sample, which is the same as the LR.

For the common benchmark where the ID dataset is CIFAR10, ER achieves the best average AUROC values across all datasets. Moreover, ER does not require fine-tuning the encoder for different test samples, which reduces processing time. All the experiments were run on a Tesla V100, ER can be much faster than LR while maintaining competitive performance. FRL is another method that can process images fast while maintaining a good performance, but the average AUROC value of FRL is worse than ER.

 For high-resolution datasets, the scale factor used for interpolation is 8 here. ER achieves the best average performance and FRL is the second best, while FRL is faster than ER considering the processing time. The processing time of methods that require image complexity drops more than other methods since it takes more time to compress the images with the size increasing. Tilt + WIM and LR are less effective on high-resolution datasets, we assume these two methods require a longer optimisation iteration for large image sizes which can slow the processing time more.

\begin{table*}[h]
\centering
\caption{AUROC results of ER and other VAE-based OOD detection methods. The last row of each table shows the number of images that each method can process per second, where a larger value is better. The bold number represents the best result.}
\begin{adjustbox}{max width=1\textwidth,center}
\begin{tabular}{ccccccccc}
\\
\toprule
            & LR\cite{xiao2020likelihood}             & LL \cite{kingma2013auto}        & IC \cite{serra2019input}        & FRL \cite{cai2023out}          & LRatio\cite{ren2019likelihood}        & Tilt \cite{flototilted}                & Tilt + WIM\cite{flototilted}    & ER (ours)    \\
\midrule
\multicolumn{9}{l}{ID dataset: CIFAR10}   \\
\midrule
MNIST       & 0.986          & 0          & 0.976       & 0.984          & 0.032         & 0.797                & 0.941        & 0.987  \\
FMNIST      & 0.976          & 0.032       & 0.987      & 0.993          & 0.335         & 0.688                & 0.939        & 0.984    \\
SVHN        & 0.912          & 0.209      & 0.938       & 0.854          & 0.732         & 0.143                & 0.787        & 0.954             \\
LSUN        & 0.606          & 0.833      & 0.348       & 0.449          & 0.508         & 0.933                & 0.650        & 0.654    \\
CelebA      & 0.738          & 0.676      & 0.310      & 0.608          & 0.404         & 0.877                 & 0.846        & 0.621             \\
Noise       & 0.994          & 1          & 0.042      & 0.925          & 0.851         & 1                     & 0.797        & 1       \\
Constant    & 0.974         & 0.015       & 1          & 1              & 0.902         & 0                     & 1            & 1     \\
\midrule
Average     & 0.885      & 0.395      & 0.657      & 0.830          & 0.538                 & 0.634             & 0.851       & \textbf{0.886}     \\
\midrule
Images$/s$  & 1.2        & 268.8     & 245.1     & 227.9            & 190.1                 & \textbf{ 295.9}   & 13.8        & 186.9     \\
\midrule
ID dataset: CelebA    \\
\midrule
iNaturalist      & 0.808       & 0.993       & 0.955    & 0.995          & 0.969         & 0.845                 & 0.650         & 0.985\\
Places           & 0.928       & 0.933       & 0.976    & 0.991          & 0.847         & 0.922                 & 0.734         & 0.993\\
SUN              & 0.929       & 0.945       & 0.959    & 0.987          & 0.884         & 0.953                 & 0.764         & 0.995\\
Textures         & 0.842       & 0.938       & 0.918    & 0.965          & 0.891         & 0.775                 & 0.606         & 0.970 \\
\midrule
Average         & 0.877       & 0.952       & 0.952     & 0.984         & 0.898         & 0.874                 & 0.689         & \textbf{0.986} \\
\midrule
Images$/s$       & 0.8         & 108.7       & 51.0      & 31.4         & 69.4           &\textbf{158.7}         & 8.4           & 23.3 \\
\bottomrule
\end{tabular}
\label{tab:compare}
\end{adjustbox}
\end{table*}

\subsection{Ablation Studies}\label{ab}
To figure out the effect of different choices of our proposed method, we conduct comprehensive ablation studies on CIFAR10 and show results in the following.

\subsubsection{Different lossy inputs}
\begin{table}[h]
\centering
\caption{AUROC results of ER with different lossy inputs, the ID dataset is CIFAR10.}
\begin{tabular}{ccccccccc}
\\
\toprule
         & Nearest        & Bilinear       & 75\% Mask & $\beta$ = 10  \\
\midrule
Average  & \textbf{0.886}   & 0.696        & 0.865      & 0.821        \\
\bottomrule
\end{tabular}
\label{tab:lossy-fashion}
\end{table}
The performance of ER depends on the operation used to obtain the lossy inputs. Therefore, we investigate four operations and report the corresponding AUROC results: nearest neighbour interpolation and bilinear interpolation with scale factors equal to 4, mask operation with $75\%$ mask ratio and the reconstructions of $\beta$-VAE with $\beta$ = 10. For both four operations, we choose the setup that can achieve the best results for a fair comparison.

The results are shown in Table~\ref{tab:lossy-fashion}. The nearest neighbour interpolation produces a more lossy image compared to bilinear interpolation since it will generate more checkerboards in the image and the former interpolation method performs better on average than the latter interpolation method. The performances of mask operation and $\beta$-VAE are better than bilinear interpolation but slightly worse than nearest interpolation.

As the nearest neighbour interpolation shows the best result, we also investigate how the scale factor can affect the performance in Table~\ref{lossam}. The scale factor can be considered as a parameter to control the degree of information loss. Since the original image size is 32$\times$32, we choose the scale factor of $\{2,4,8,16\}$, and show the corresponding average AUROC results. Table~\ref{lossam} shows that the average performance reaches the peak at S = 4 and drops a lot when the S = 16. Overall, the loss of information should be kept at a level that is not too small, but it also cannot be too large since a lossy image that contains no information cannot be used to reconstruct the original image. The nearest neighbour interpolation can be a general choice to obtain lossy inputs for different ID datasets.

\begin{table}[t]
\caption{The average AUROC results for interpolation with different scale factors when the ID dataset is CIFAR10.}
\centering
\begin{tabular}{ccccc}
\\
\toprule
         & S = 2    & S = 4         & S = 8   & S = 16   \\
\midrule
Average  & 0.844       & \textbf{0.886}      & 0.846        & 0.780     \\
\bottomrule
\end{tabular}
\label{lossam}
\end{table}


\subsubsection{Different error functions}
There is more than one choice for the error function. In our model design, the reconstruction error is cross-entropy (CE). Rather than CE, we also test the performance for mean absolute error (MAE), mean squared error (MSE) and LPIPS \cite{zhang2018perceptual}. Table~\ref{tab:lossy-ds} shows that the CE, MAE and MSE can achieve close performance while the result of LPIPS is more likely to be a random guess. We assume that is due to the pre-trained model used in calculating the LPIPS. The pre-trained model has an input size of $224\times224$ which is larger than $32\times32$. There is no guarantee that the reconstruction of our VAE can achieve a better LPIPS value than the lossy input (obtained with interpolation methods) after an upsampling operation.

\begin{table}[h]
\centering
\caption{The average AUROC results of ER with different error functions, the ID dataset is CIFAR10.}
\begin{tabular}{ccccc}
\\
\toprule
         & CE                     & MAE                 & MSE        & LPIPS\\
\midrule
Average  & \textbf{0.886}         & 0.855               & 0.836      & 0.507\\
\bottomrule
\end{tabular}
\label{tab:lossy-ds}
\end{table}

\subsubsection{ER with different $\lambda$ values}
\begin{figure}[h]
\includegraphics[scale=0.5]{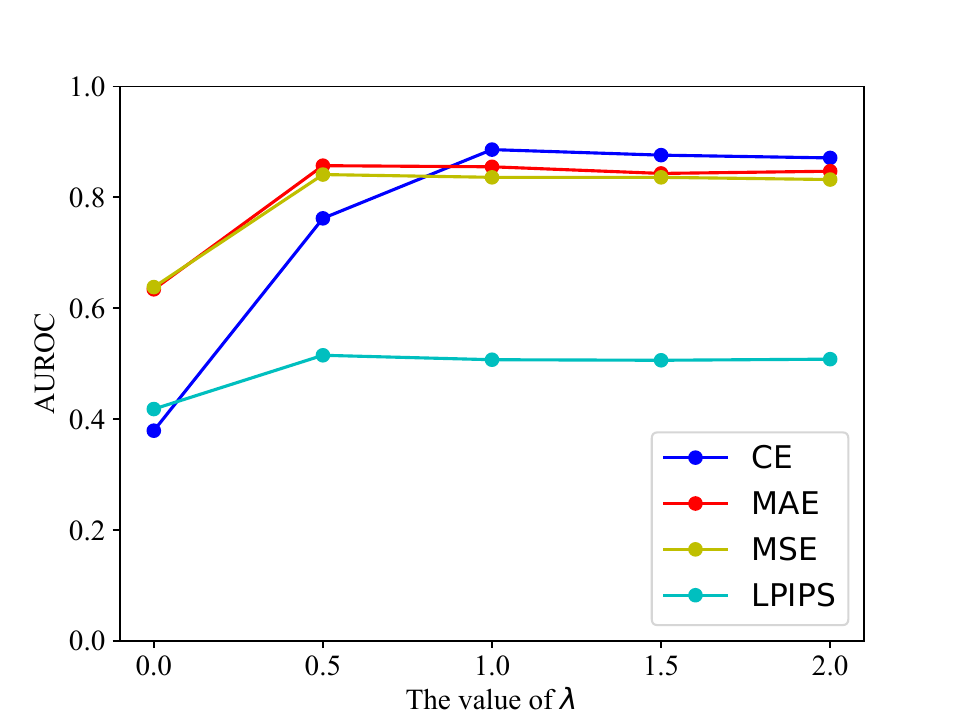}
\caption{The average AUROC results for different $\lambda$ values with different error functions when the ID dataset is CIFAR10.}
\label{losslambda}
\end{figure}

 As we adopt the input complexity as a regularisation term in the ER, we also test how the weight of the input complexity can affect the performance with different distance metric choices when the ID dataset is CIFAR10. As shown in Figure~\ref{losslambda}, the worst average performance is at $\lambda$ = 0. This is not surprising as we analysed before. CE is the worst choice when $\lambda$ = 0, this is due to that CE is a probability-based metric and the value of CE between two pixels with values 0 and 255 would be less different from 100 and 255 compared to MAE and MSE, which brings limited help for OOD detection without input complexity. All different distance metric choices can reach a stable stage after $\lambda\geq$ 0.5 except the CE gets stable after $\lambda\geq$ 1.

\subsubsection{Different pre-trained encoders}\label{difen}
As mentioned in Section~\ref{4.2}, lossy inputs reduce the gap between ID samples and OOD samples. Thus, the encoder of the VAE should bring limited effects to the results of ER as long as the decoder can reconstruct the original images by feeding lossy inputs. We investigate this assumption by using a two-stage training scheme. With different pre-trained VAEs, we freeze the encoder and only optimise the corresponding decoder to reconstruct the original image by feeding a lossy image. We test three different kinds of pre-trained VAE: a vanilla VAE, a $\beta$-VAE with $\beta=10$ and a VAE trained with lossy images as inputs and targets. We also show the results of using a vanilla VAE without second training. Table~\ref{tab:representation} shows the detailed results for pre-trained encoders for CIFAR10. ER$_{dir}$ is the only setting that cannot reconstruct the original images with lossy inputs and it shows the worst results while other settings show similar results. The results support our point that ER can perform well with different encoders when the corresponding decoder can reconstruct the original images.
\begin{table}[t]
\caption{AUROC results of ER with different pre-trained encoders. ER represents the original setting, ER$_{dir}$ represents using a vanilla VAE directly, ER$_{van}$ represents optimising the decoder of a vanilla VAE, ER$_{van}$ represents optimising the decoder of a $\beta$-VAE with $\beta$ = 10, and ER$_{lossy}$ represents optimising the decoder of a VAE trained with lossy images as both inputs and targets.}
\begin{tabular}{cccccc}
\\
\toprule
         & ER                & ER$_{dir}$   & ER$_{van}$        & ER$_{\beta}$  & ER$_{lossy}$  \\
\midrule
MNIST    & 0.987             & 0.403           & 0.983        & 0.954        & 0.976  \\
FMNIST   & 0.984             & 0.556           & 0.987        & 0.969        & 0.981  \\
SVHN     & 0.954             & 0.505           & 0.946        & 0.942        & 0.952   \\
LSUN     & 0.654             & 0.679           & 0.617        & 0.556        & 0.623   \\
CelebA   & 0.621             & 0.457           & 0.576        & 0.626        & 0.610  \\
Noise    & 1                 & 0.647           & 0.997        & 1            & 1         \\
Constant & 1                 & 0.417           & 0.998        & 0.998        & 0.994     \\
\midrule
Average  & \textbf{0.886}             & 0.523           & 0.872        & 0.864        & 0.877  \\
\bottomrule
\end{tabular}

\label{tab:representation}
\end{table}

\section{Conclusion and future work}
OOD detection is crucial in deploying real-world deep learning applications, and developing efficient algorithms to address this issue is urgent. In this paper, we proposed a VAE-based unsupervised OOD score that achieves the SOTA average AUROC results without fine-tuning. Concurrently, our method can also maintain a faster processing time compared to other methods such as LR. With comprehensive experimental results, we suggest that the nearest interpolation can be a general choice as the lossy operation. However, it is also possible that the algorithm may not detect malicious attacks or anomalous cases as there are no perfect solutions. Therefore, supervision is still required at present.

ER can potentially be used in other generative models such as diffusion models \cite{song2021denoising}. Although the encoding process of the diffusion model is adding noise to images rather than encoding lossy images. We leave these investigations as future works.

\paragraph{}\textbf{Competing Interests} The authors declare no competing interests.
\vspace{-0.5cm}
\paragraph{}\textbf{Authors contribution statement} \\
Zezhen Zeng carried out the conceptualization, investigation, methodology, software, validation, visualization and writing. Bin Liu carried out the supervision and writing - review $\&$ editing.
\vspace{-0.5cm}
\paragraph{}\textbf{Ethical and informed consent for data used} There are no human subjects or private data in this paper and informed consent is not applicable.
\vspace{-0.5cm}

\paragraph{}\textbf{Data availability and access} All data are available online. MNIST is available
in \cite{lecun1998mnist}. CIFAR10 is available in \cite{krizhevsky2009learning}. KMNIST is in \cite{clanuwat2018deep}. FashionMNIST is available in \cite{xiao2017fashion}. CelebA is in \cite{liu2015faceattributes}. Noise and Constant are synthetic datasets detailed in \cite{xiao2020likelihood}.

\bibliography{ecs}
\bibliographystyle{IEEEbib}


\end{document}